%% file: paper.tex
\documentclass[review]{elsarticle}

\usepackage{hyperref}
\usepackage{booktabs}
\usepackage{amsmath,amsfonts}
\usepackage{multirow}

\DeclareMathOperator*{\argmax}{\arg\!\max}

\journal{ArXiv}

\begin{document}

\begin{frontmatter}

\title{Fast Ensemble Learning Using Adversarially-Generated Restricted Boltzmann Machines}

\author{Gustavo H. de Rosa\corref{c}}
\cortext[c]{Corresponding author}
\ead{gustavo.rosa@unesp.br}

\author{Mateus Roder}
\ead{mateus.roder@unesp.br}

\author{Jo\~ao Paulo Papa}
\ead{joao.papa@unesp.br}

\address{Department of Computing \\S\~ao Paulo State University \\ Bauru, Brazil
}

\begin{abstract}
Machine Learning has been applied in a wide range of tasks throughout the last years, ranging from image classification to autonomous driving and natural language processing. Restricted Boltzmann Machine (RBM) has received recent attention and relies on an energy-based structure to model data probability distributions. Notwithstanding, such a technique is susceptible to adversarial manipulation, i.e., slightly or profoundly modified data. An alternative to overcome the adversarial problem lies in the Generative Adversarial Networks (GAN), capable of modeling data distributions and generating adversarial data that resemble the original ones. Therefore, this work proposes to artificially generate RBMs using Adversarial Learning, where pre-trained weight matrices serve as the GAN inputs. Furthermore, it proposes to sample copious amounts of matrices and combine them into ensembles, alleviating the burden of training new models'. Experimental results demonstrate the suitability of the proposed approach under image reconstruction and image classification tasks, and describe how artificial-based ensembles are alternatives to pre-training vast amounts of RBMs.
\end{abstract}


\end{frontmatter}

\input{introduction.tex}
\input{background.tex}
\input{proposed.tex}
\input{methodology.tex}
\input{experiment.tex}
\input{conclusion.tex}

\section*{Conflicts of Interest}

The authors declare that there are no conflicts of interest.

\section*{Acknowledgments}

The authors are grateful to S\~ao Paulo Research Foundation (FAPESP) grant \#2019/02205-5.

\bibliography{references}

\end{document}

%% file: introduction.tex
\section{Introduction}
\label{s.introduction}

Artificial Intelligence (AI) has become one of the most cherished areas throughout the last decades, essentially due to its capability in supporting humans in decision-making tasks and automatizing repetitive jobs~\cite{Bishop:06}. The key-point for AI's remarkable results lies in a subarea commonly known as Machine Learning (ML). Such area is in charge of researching and developing algorithms that are proficient in solving tasks through models that can learn from examples, such as image classification~\cite{Wang:17}, object recognition~\cite{Wang:16}, autonomous driving~\cite{Sallab:17}, and natural language processing~\cite{Deng:18}, among others.

An algorithm that has received spotlights is the Restricted Boltzmann Machine (RBM)~\cite{Hinton:02}, which is an energy-based architecture that attempts to represent the data distribution relying upon physical and probabilistic theories. It can also reconstruct data in a learned latent space, acting as an auto encoder-decoder capable of extracting new features from the original data~\cite{Srivastava:12}. Nevertheless, such a network, along with other ML algorithms, often suffers when reconstructing manipulated data, e.g., slightly or profoundly modified information, being unfeasible when employed under real-world scenarios\footnote{Usually, real-world data have higher variance than experimental data and often results in poorly performances.}~\cite{Szegedy:13}.

A noteworthy perspective that attempts to overcome such a problem is Adversarial Learning (AL), which introduces adversarial data (noisy data) to a network's training, aiding in its ability to deal with high variance information~\cite{Laskov:10}. Notwithstanding, it is not straightforward to produce noisy data and feed to the network, as too much noise will impact the architecture's performance, while too little noise will not help the network in learning distinct patterns~\cite{Li:17}. Goodfellow et al.~\cite{Goodfellow:14} introduced the Generative Adversarial Network (GAN) as a solution to AL's problem, where a two-network system competes in a zero-sum approach. In summary, a GAN is composed of both discriminative and generative networks, commonly known as the discriminator and the generator. The generator is in charge of producing fake data, while the discriminator estimates the probability of the fake data being real.

One can observe in the literature that several works have successfully applied GANs in adversarial data generation, specifically images. For instance, Ledig et al.~\cite{Ledig:17} introduced the Super-Resolution Generative Adversarial Network (SRGAN), capable of inferring photo-realistic images from upscaling factors by employing a perceptual loss function that can guide the network in recovering textures from downsampled images. Choi et al.~\cite{Choi:18} proposed the StarGAN, which is a scalable approach for performing image-to-image translations over multiple domains. They proposed a unified architecture that simultaneously trains several datasets with distinct domains, leading to a superior quality of the translated images. Moreover, Ma et al.~\cite{Ma:19} presented a novel framework that fuses features from infrared and visible images, known as FusionGAN. Essentially, their generator creates fused images with high infrared intensities and additional visible features, while their discriminator forces the images to have more details than the visible ones.

\begin{sloppypar}
Notwithstanding, only a few works prosperously employed GANs and energy-based models, e.g., RBMs. Zhao et al.~\cite{Zhao:16} presented an Energy-based Generative Adversarial Network (EBGAN) to generate high-resolution images, which uses an energy-based discriminator that assigns high energy values to generated samples and a generator that produces contrastive samples with minimal energy values, allowing a more stable training behavior than traditional GANs. Luo et al.~\cite{Luo:18} proposed an alternative training procedure for ClassRBMs in the context of image classification, denoted as Generative Objective and Discriminative Objective (ANGD). Essentially, they used GAN-based training concepts, where the parameters are firstly updated according to the generator, followed by the discriminator's updates. Furthermore, Zhang et al.~\cite{Zhang:19} introduced the Adversarial Restricted Boltzmann Machine (ARBM) in high-quality color image generation, which minimizes an adversarial loss between data distribution and the model's distribution without resorting to explicit gradients. Nonetheless, to the best of the authors' knowledge, no single work attempts to use GANs to generate artificial RBMs.
\end{sloppypar}

Therefore, this work proposes to pre-train RBMs in a training set and use their weight matrices as the input data for GANs. In other words, the idea is to learn feasible representations of the input weight matrices and further generate artificial matrices, where each generated matrix will be used to construct a new RBM and evaluate its performance in a testing set. Additionally, this paper proposes to sample vasts amounts of artificial-based RBMs and construct heterogeneous ensembles to alleviate the burden of pre-training plentiful amounts of new RBMs. Therefore, this work has three main contributions:

\begin{itemize}
\item Introduce a methodology to generate artificial RBMs;
\item Apply artificially-generated RBMs in the context of image reconstruction and classification;
\item Construct heterogeneous ensembles to lessen the computational load of pre-training new models.
\end{itemize}

The remainder of the paper is organized as follows. Section~\ref{s.background} introduces the main concepts related to Restricted Boltzmann Machines and Generative Adversarial Networks. Section~\ref{s.proposed} describes the proposed approach considering the Adversarially-Generated RBMs, while Section~\ref{s.methodology} details the methodology. Section~\ref{s.experiments} presents the experimental results and some insightful discussions. Finally, Section~\ref{s.conclusion} states conclusions and future works.

%% file: background.tex
\section{Theoretical Background}
\label{s.background}

In this section, we present a brief explanation of Restricted Boltzmann Machines and Generative Adversarial Networks.

\subsection{Restricted Boltzmann Machines}
\label{ss.rbm}

Restricted Boltzmann Machines are stochastic-based neural networks and inspired by physical concepts, such as energy and entropy. Usually, RBMs are trained through unsupervised algorithms and applied in image-based tasks, such as collaborative filtering, feature extraction, image reconstruction, and classification.

\begin{figure}[!ht]
    \centering
    \includegraphics[scale=0.65]{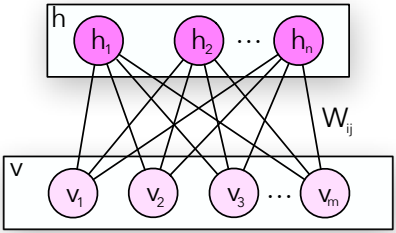}
    \caption{Standard RBM architecture.}
    \label{f.rbm}
\end{figure}

An RBM comprises a visible layer $\mathbf{v}$ composed by $m$ units, which deals directly with the input data, and a hidden layer $\mathbf{h}$ constituted of $n$ units, which is in charge of learning features and the input data's probabilistic distribution. Also, let $\mathbf{W}_{m \times n}$ be the matrix that models the weights between visible and hidden units, where $w_{ij}$ stands for the connection between visible unit $v_i$ and hidden unit $h_j$. Figure~\ref{f.rbm} illustrates a standard RBM architecture\footnote{Note that an RBM does not have connections between units in the same layer.}.

A deriving model from the RBM, often known as Bernoulli-Bernoulli RBM (BBRBM), assumes that all units in layers $\mathbf{v}$ and $\mathbf{h}$ are binary and sampled from a Bernoulli distribution~\cite{Hinton:12}, i.e., $\mathbf{v}\in\{0,1\}^m$ and  $\mathbf{h}\in\{0,1\}^n$. Under such an assumption, Equation~\ref{e.energy_bbrbm} describes the energy function of a BBRBM:

\begin{equation}
\label{e.energy_bbrbm}
E(\mathbf{v},\mathbf{h})=-\sum_{i=1}^ma_iv_i-\sum_{j=1}^nb_jh_j-\sum_{i=1}^m\sum_{j=1}^nv_ih_jw_{ij},
\end{equation}
where $\mathbf{a}$ and $\mathbf{b}$ stand for the visible and hidden units biases, respectively.

Additionally, the joint probability of a given configuration $(\mathbf{v},\mathbf{h})$ is modeled by Equation~\ref{e.probability_configuration} , as follows:

\begin{equation}
\label{e.probability_configuration}
P(\mathbf{v},\mathbf{h})=\frac{e^{-E(\mathbf{v},\mathbf{h})}}{Z},
\end{equation}
where $Z$ stands for the partition function, responsible for normalizing the probability over all possible visible and hidden units configurations. 

Moreover, the probability of a visible (input) vector is descrived by Equation~\ref{e.probability_configuration2}, as follows:

\begin{equation}
\label{e.probability_configuration2}
P(\mathbf{v})=\frac{\displaystyle\sum_{\mathbf{h}}e^{-E(\mathbf{v},\mathbf{h})}}{Z}.
\end{equation}

One can perceive that a BBRBM is a bipartite graph, hence, it allows information to flow from visible to hidden units, and vice-versa. Therefore, it is possible to employ Equations~\ref{e.pv} and~\ref{e.ph} to formulate mutually independent activations for both visible and hidden units, as follows:

\begin{equation}
\label{e.pv}
 P(\mathbf{v}|\mathbf{h})=\prod_{i=1}^mP(v_i|\mathbf{h})
\end{equation}
and

\begin{equation}
\label{e.ph}
 P(\mathbf{h}|\mathbf{v})=\prod_{j=1}^nP(h_j|\mathbf{v}),
\end{equation}
where $P(\mathbf{v}|\mathbf{h})$ and $P(\mathbf{h}|\mathbf{v})$ are the probability of the visible layer given the hidden states and the probabily of the hidden layer given the visible states, respectively.

Moreover, from Equations~\ref{e.pv} and~\ref{e.ph}, one can achieve the probability of activating a single visible unit $i$ given hidden states and the probability of achieving a single hidden unit $j$ given visible states. Such activations are described by Equations~\ref{e.probv} and~\ref{e.probh}, as follows:

\begin{equation}
\label{e.probv}
P(v_i=1|\mathbf{h})=\sigma\left(\sum_{j=1}^nw_{ij}h_j+a_i\right)
\end{equation}
and

\begin{equation}
\label{e.probh}
P(h_j=1|\mathbf{v})=\sigma\left(\sum_{i=1}^mw_{ij}v_i+b_j\right),
\end{equation}
where $\sigma(\cdot)$ is the logistic-sigmoid function.

The training algorithm of a BBRBM learns a set of parameters $\theta=(\mathbf{W}, \mathbf{a}, \mathbf{b})$ through an optimization problem, which aims at maximizing the product of probabilities derived from a training set ${\cal D}$. Equation~\ref{e.rbm_opt} models such a problem, as follows:

\begin{equation}
\label{e.rbm_opt}
    \argmax_{\Theta}\prod_{\mathbf{d}\in{\cal D}}P(\mathbf{d}).
\end{equation}

An alternative to solve such optimization is by applying the negative of the logarithm function, known as Negative Log-Likelihood (NLL). It represents an approximation of both reconstructed data and original data distributions. A better alternative proposed by Hinton et al.~\cite{Hinton:02}, known as Contrastive Divergence (CD), uses the training data as the initial visible units and the Gibbs sampling methods to infer the hidden and reconstructed visible layers.

Finally, one can apply derivatives and formulate the parameters update rule, described by Equations~\ref{e.updateW},~\ref{e.updatea} and~\ref{e.updateb}, as follows:

\begin{equation}
\label{e.updateW}
\mathbf{W}^{s+1} = \mathbf{W}^s + \eta(\mathbf{v}P(\mathbf{h}|\mathbf{{v}}) - \mathbf{\tilde{v}}P(\mathbf{\tilde{h}}|{\mathbf{\tilde{v}}})),
\end{equation}

\begin{equation}
\label{e.updatea}
\mathbf{a}^{s+1} = \mathbf{a}^s + (\mathbf{v} - \mathbf{\tilde{v}})
\end{equation}
and

\begin{equation}
\label{e.updateb}
\mathbf{b}^{s+1} = \mathbf{b}^{s} + (P(\mathbf{h}|\mathbf{v}) - P(\mathbf{\tilde{h}}|\mathbf{\tilde{v}})),
\end{equation}
where $s$ stands for the current epoch, $\eta$ is the learning rate, $\mathbf{\tilde{v}}$ stands for the reconstruction of the visible layer given $\mathbf{h}$, and $\mathbf{\tilde{h}}$ is an approximation of the hidden vector $\mathbf{h}$ given $\mathbf{\tilde{v}}$.

\subsection{Generative Adversarial Networks}
\label{ss.gan}

\begin{figure}[!ht]
    \centering
    \includegraphics[scale=0.5]{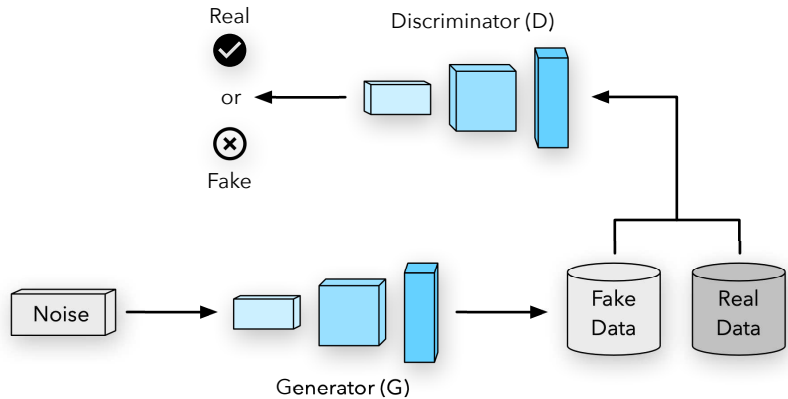}
    \caption{Standard architecture of a Generative Adversarial Network.}
    \label{f.gan}
\end{figure}

Goodfellow et al.~\cite{Goodfellow:14} introduced the Generative Adversarial Networks, essentially composed of discriminative and generative networks that compete among themselves in a zero-sum game. The discriminative part estimates the probability of a fake sample being a real one, while the generative part produces the fake samples. Figure~\ref{f.gan} illustrates an example of a standard Generative Adversarial Network.

Regarding its training algorithm, the discriminative network $D$ is trained to maximize the probability of classifying data as real images, regardless of them being real or generated. Concurrently, the generative network $G$ is trained to minimize the divergence between real and fake data distributions, i.e., $log(1 - D(G(\mathbf{\zeta}))$. As mentioned before, the two neural networks compete between themselves in a zero-sum game, trying to achieve an equilibrium, which is represented by Equation~\ref{e.gan_min_max}:

\begin{equation}
\label{e.gan_min_max}
\min_G \max_D C(D,G) = \mathbb{E}_{\mathbf{x}}[logD(\mathbf{x})] + \mathbb{E}_{\mathbf{\zeta}}[log(1-D(G(\mathbf{\zeta}))],
\end{equation}
where $C(D,G)$ stands for the loss function to be minimized, $D(\mathbf{x})$ stands for the estimated probability of a real sample $\mathbf{x}$ being real, $\mathbb{E}_{\mathbf{x}}$ is the mathematical expectancy over all samples from the real data set $\mathcal X$, $G(\mathbf{\zeta})$ stands for the generated data given the noise vector $\mathbf{\zeta}$, $D(G(\mathbf{\zeta}))$ is the estimated probability of a fake sample $G(\mathbf{\zeta})$ being real, and $\mathbb{E}_{\mathbf{\zeta}}$ is the mathematical expectancy over all random generator inputs, i.e., the expected value over all fake samples generated by $G$.

Notwithstanding, Equation~\ref{e.gan_min_max} imposes a problem where GANs can get trapped in local optimums when the discriminator faces an easy task. Essentially, at the early training iterations, when $G$ still does not know how to generate adequate samples, $D$ might reject the generated samples with high probabilities\footnote{Such a problem saturates the function $log(1-D(G(\mathbf{\zeta}))$.}, leading to inadequate training. Therefore, an alternative to overcome such a problem is to train $G$ to maximize $log(D(G(\mathbf{\zeta}))$, improving initial gradients and mitigating the possibility of getting trapped in local optimums.

%% file: proposed.tex
\section{Proposed Approach}
\label{s.proposed}

\begin{figure}[!ht]
    \centering
    \includegraphics[scale=0.6]{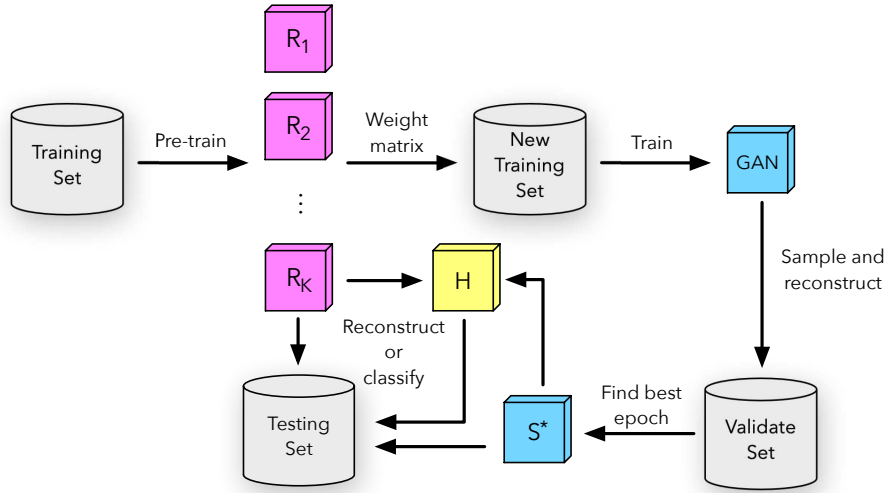}
    \caption{Proposed approach pipeline.}
    \label{f.proposed}
\end{figure}

This section provides a more in-depth explanation of how the proposed approach works, divided into three parts: pre-training, adversarial learning, and sampling. Additionally, we describe how to construct artificially-based ensembles. Figure~\ref{f.proposed} illustrates the proposed approach pipeline to provide a more precise visualization.

\subsection{Pre-Training RBMs}
\label{ss.pre_train_rbms}

Let $t$ be a sample that belongs to a training set $\mathcal T$, $K$ be a set of RBMs, as well as $\mathbf{W}_k$ be the weight matrix from a given RBM $R_k$. Initially, every RBM is trained on every sample from the training set $\mathcal T$, resulting in $K$ pre-trained RBMs and, hence, $K$ pre-trained weight matrices. Further, the $K$ weight matrices are concatenated into a new training set, denoted as $\mathcal T_A$, which is used to feed the Adversarial Learning training. Equations~\ref{e.rbm_weight} and~\ref{e.rbm_concat} formulate such a process, as follows:

\begin{equation}
\label{e.rbm_weight}
\mathbf{W}_k = R_k(t) \mid \forall t \in \mathcal T \text{, where} \: k \in \{1, 2, \ldots, K\}
\end{equation}
and

\begin{equation}
\label{e.rbm_concat}
\mathcal T_A = [\mathbf{W}_1, \mathbf{W}_2, \ldots, \mathbf{W}_K]^T.
\end{equation}

Therefore, one can see that the intended approach aims to use pre-trained weight matrices as the GAN's input to learn their patterns and sample new matrices.

\subsection{Adversarial Learning}
\label{ss.train_adv}

After performing the RBMs pre-training, it is possible to train a GAN with the new training set $\mathcal T_A$. In other words, the idea is to use the pre-trained weight matrices as the GAN's input to train its generator and learn how to generate artificial matrices. Nevertheless, due to the difficulty in establishing an equilibrium when training GANs, we opted to employ an additional validation set, denoted as $\mathcal V$, which is used to assess whether an artificial matrix is suitable.

Before the GAN's training process, we randomly select an original RBM and resets its biases, e.g., $R_r$, where $r \in [1, K]$ is a randomly sampled integer. Then, we sample a new weight matrix $\mathbf{\tilde{W}}_r$ from its generator for every training epoch $d$, such that $d \in \{1, 2, \ldots, D\}$ and $D$ is the maximum number of epochs. Such a procedure is described by Equation~\ref{e.gan_sample}:

\begin{equation}
\label{e.gan_sample}
\mathbf{\tilde{W}}_r = G(\mathbf{\zeta}_d) \mid \text{where} \: d \in \{1, 2, \ldots, D\}.
\end{equation}

Afterward, $\mathbf{\tilde{W}}_r$ is replaced as $R_r$ weight matrix and reconstructed over the validation set $\mathcal V$. Finally, after reconstructing $D$ weight matrices, we select the epoch that achieved the lowest mean squared error (MSE) as our final GAN, denoted as $S^{\ast}$.

\subsection{Sampling New RBMs}
\label{ss.adv_rbm}

Finally, one can now sample $L$ weight matrices from $S^{\ast}$ and artificially-create $L$ new RBMs. Furthermore, to verify whether the artificially generated RBMs are suitable, they are reconstructed over testing set $\mathcal Z$ and compared against the $K$ original reconstructed RBMs. Note that the proposed approach is extensible to the task of image classification, where features are extracted from an RBM's hidden layer and fed to a classifier.

\subsection{Constructing Ensembles}
\label{ss.adv_ensemble}

Alternatively, with a pre-trained GAN in hands, one can sample a vast amount of artificial RBMs and compose a heterogeneous ensemble along with the original RBMs. Let $H$ be an ensemble composed of $K$ original and $L$ artificial RBMs. Additionally, let $\hat{y}_i$, where $i \in \{1, 2, \ldots, K+L\}$ be the prediction of the ith $i$ RBM over a testing set $\mathcal Z$. One can construct an array of predicted labels, denoted as $\mathbf{y}(z)$, which holds the predictions of all $K$ and $L$ RBMs over sample $z \in \mathcal Z$. Finally, the heterogeneous ensemble combines all predictions using majority voting, as follows:

\begin{equation}
    \label{e.weighted_voting}
    H(z) = \argmax_{j \in \{1, 2, \ldots, C\}} \sum_{i=1}^{K+L}\hat{y}_{i,j}(z),
\end{equation}
where $C$ stands for the number of classes.

%% file: methodology.tex
\section{Methodology}
\label{s.methodology}

In this section, we discuss the methodology used to conduct this work, the employed dataset, the evaluation tasks, and the experimental setup\footnote{The source code is available at \url{https://github.com/gugarosa/synthetic_rbms}.}.

\subsection{Evaluation Tasks}
\label{ss.tasks}

The proposed approach is evaluated over distinct tasks, such as image reconstruction and image classification. Regarding image reconstruction, the objective function is guided by the mean square error, which stands for the error between the original and the reconstructed images. Furthermore, concerning the image classification task, we opted to use an additional classifier, such as the Support Vector Machine~\cite{Chang:11} with radial kernel and without parameter optimization, instead of a Discriminative Restricted Boltzmann Machine~\cite{Larochelle:08}. Therefore, instead of reconstructing samples with the original or artificial RBMs, we use them as feature extractors, where the original samples are passed to the hidden layer, and their activations are passed to the classifier. Afterward, the classifier outputs an accuracy value, which stands for the percentage of correct labels assigned by a classifier and guides the proposed task. Additionally, we opted to construct original and artificial RBMs ensembles to provide a more in-depth comparison and enhance the proposed approach.

\subsection{Dataset}
\label{ss.dataset}

We opted to consider only one toy-dataset to conduct the proposed experiments as we wanted to explore the suitability and how the proposed method would behave under a theoretical scenario. The dataset is the well-known MNIST\footnote{\url{http://yann.lecun.com/exdb/mnist}}~\cite{Lecun:98}, which is composed of a set of $28 \times 28$ grayscale images of handwritten digits, containing a training set with $60,000$ images from digits `0'-`9', as well as a testing set with $10,000$ images. Additionally, as our proposed approach uses a validation set, we opted to split the training set into two new sets: (i) $48,000$ training images and (ii) $12,000$ validation images.

\subsection{Experimental Setup}
\label{ss.setup}

Table~\ref{t.rbm_setup} describes the experimental setup used to pre-train the RBMs, implemented using Learnergy~\cite{Roder:20} library. Note that as we are trying to understand whether the proposed approach is suitable or not in the evaluated context, we opted to use the simplest version of RBM, i.e., without any regularization, such as momentum, weight decay, and temperature.

\begin{table}[!ht]
    \renewcommand{\arraystretch}{1.25}
    \centering
    \caption{RBM pre-training hyperparameter configuration.}
    \label{t.rbm_setup}
    \begin{tabular}{lc}
        \toprule
        \textbf{Hyperparameter} & \textbf{Value} \\
        \midrule
        $m$ (number of visible units) & $784$ \\
        $n$ (number of hidden units) & $[32, 64, 128]$ \\
        $steps$ (number of CD steps) & $1$ \\
        $\eta$ (learning rate) & $0.1$ \\
        $bs$ (batch size) & $128$ \\
        $epochs$ (number of training epochs) & $10$ \\
        \bottomrule
    \end{tabular}
\end{table}

Table~\ref{t.gan_setup} describes the experimental setup used to train the GANs, implemented using NALP\footnote{\url{https://github.com/gugarosa/nalp}} library. Before the real experimentation, we opted to conduct a grid-search procedure to find adequate values for the number of noise dimensions, the batch size, and training epochs. Additionally, we opted to use a low learning rate due to the high number of training epochs and more stable convergence. Regarding the GAN architecture, we used a linear architecture composed of three down-samplings ($512, 256, 128$ units), ReLU activations ($0.01$ threshold), and a single-unit output layer for the discriminator, as well as three up-samplings ($128, 256, 512$), ReLU activations ($0.01$ threshold) and an output layer with a hyperbolic tangent activation. Considering the ensemble construction in image classification, we opted to use its most straightforward approach, i.e., majority voting, which employs $K$ models to create an array of $K$ predictions, gathering the most voted prediction final label.

\begin{table}[!ht]
    \renewcommand{\arraystretch}{1.25}
    \centering
    \caption{GAN pre-training hyperparameter configuration.}
    \label{t.gan_setup}
    \begin{tabular}{lc}
        \toprule
        \textbf{Hyperparameter} & \textbf{Value} \\
        \midrule
        $n_z$ (number of noise dimensions) & $10000$ \\
        $\eta_D$ (discriminator learning rate) & $0.0001$ \\
        $\eta_G$ (generator learning rate) & $0.0001$ \\
        $K$ (number of pre-trained RBMs) & $[32, 64, 128, 256, 512]$ \\
        $bs$ (batch size) & $\frac{K}{16}$ \\
        $E$ (number of training epochs) & $4000$ \\
        \bottomrule
    \end{tabular}
\end{table}

%% file: experiment.tex
\section{Experimental Results}
\label{s.experiments}

This section presents the experimental results concerning the tasks of image reconstruction and image classification. 

\subsection{Image Reconstruction}
\label{ss.experiments_rec}

Table~\ref{t.rbm_rec_exp} describes the mean reconstruction errors and their standard deviations over the MNIST testing set\footnote{Note that we opted to sample $K$ weights, which is the same amount of pre-trained RBMs.}, concerning the original RBMs ($R$) and artificially-generated ones ($S^{\ast}$). According to the Wilcoxon signed-rank test with $5\%$ significance, the bolded cells are statistically equivalent, and the underlined ones are the lowest mean reconstruction errors.

\begin{table}[!ht]
	\renewcommand{\arraystretch}{1.25}
	\caption{Mean reconstruction error and standard deviation over MNIST testing set.}
	\label{t.rbm_rec_exp}
	\centering
	\begin{tabular}{cccc}
	\toprule
	\textbf{Hidden Units} & \textbf{Number of RBMs} & $\mathbf{R}$ & $\mathbf{S^{\ast}}$ \\
	\midrule
	\multirow{5}{*}{$32$} & $32$ & $\underline{\mathbf{88.173 \pm 1.053}}$ & $110.779 \pm 4.120$ \\
	& $64$ & $\underline{\mathbf{88.418 \pm 1.134}}$ & $103.484 \pm 1.015$ \\
	& $128$ & $\underline{\mathbf{88.607 \pm 1.091}}$ & $120.138 \pm 10.217$ \\
	& $256$ & $\underline{\mathbf{88.639 \pm 1.040}}$ & $121.924 \pm 13.171$ \\
	& $512$ & $\underline{\mathbf{88.662 \pm 1.063}}$ & $152.059 \pm 17.867$ \\
	\midrule
	\multirow{5}{*}{$64$} & $32$ & $\underline{\mathbf{75.219 \pm 0.881}}$ & $92.628 \pm 6.344$ \\
	& $64$ & $\underline{\mathbf{75.351 \pm 0.969}}$ & $90.071 \pm 2.443$ \\
	& $128$ & $\underline{\mathbf{75.393 \pm 0.910}}$ & $102.536 \pm 7.133$ \\
	& $256$ & $\underline{\mathbf{75.326 \pm 0.941}}$ & $128.106 \pm 22.970$ \\
	& $512$ & $\underline{\mathbf{75.274 \pm 0.887}}$ & $137.621 \pm 29.069$ \\
	\midrule
	\multirow{5}{*}{$128$} & $32$ & $\underline{\mathbf{64.768 \pm 0.596}}$ & $72.607 \pm 5.870$ \\
	& $64$ & $\underline{\mathbf{64.748 \pm 0.616}}$ & $76.660 \pm 7.823$ \\
	& $128$ & $\underline{\mathbf{64.756 \pm 0.643}}$ & $71.581 \pm 5.830$ \\
	& $256$ & $\underline{\mathbf{64.776 \pm 0.677}}$ & $95.401 \pm 18.083$ \\
	& $512$ & $\underline{\mathbf{64.725 \pm 0.683}}$ & $106.867 \pm 30.411$ \\
	\bottomrule
	\end{tabular}
\end{table}

Artificial RBMs are generated from noise and may loose performance when sampled from large sets of input data. A considerable $K$ value indicates that $K$ artificial RBMs were sampled from a GAN trained with $K$ RBMs, hence providing a diversity factor and increasing their reconstruction errors and standard deviations. One can perceive that every $K=\{256,512\}$ artificial RBMs achieved the highest reconstruction errors and standard deviations, indicating a performance loss and a worse reconstruction when compared to $K=\{32,64,128\}$ experiments. On the other hand, such behavior can not be detected in standard RBMs as they are trained from scratch with the same data and often achieves similar results.

Another interesting point is that many hidden units ($n=128$) decreased the gap between $R$ and $S^{\ast}$ reconstruction errors, especially when used within lower amounts of $K$, e.g., $K=\{32,64,128\}$. As a higher number of hidden units provides more consistent training and better reconstructions, it is more feasible that GANs trained with these RBMs performs better samplings, which results in lower reconstruction errors.

\subsubsection{Analyzing GANs Convergence}
\label{sss.experiments_rec_convergence}

\begin{figure}[!ht]
	\centering
	\includegraphics[scale=0.5]{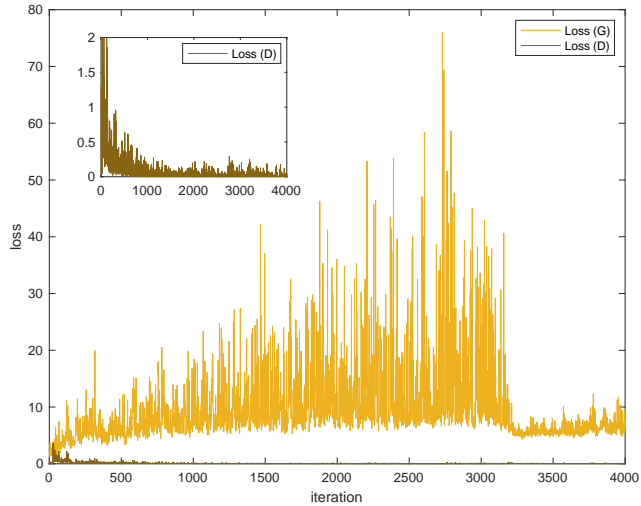}
	\caption{Training set losses convergence by GANs with $n=32$ and $K=128$ pre-trained inputs.}
    \label{f.train_loss_32}
\end{figure}

\begin{figure}[!ht]
	\centering
	\includegraphics[scale=0.5]{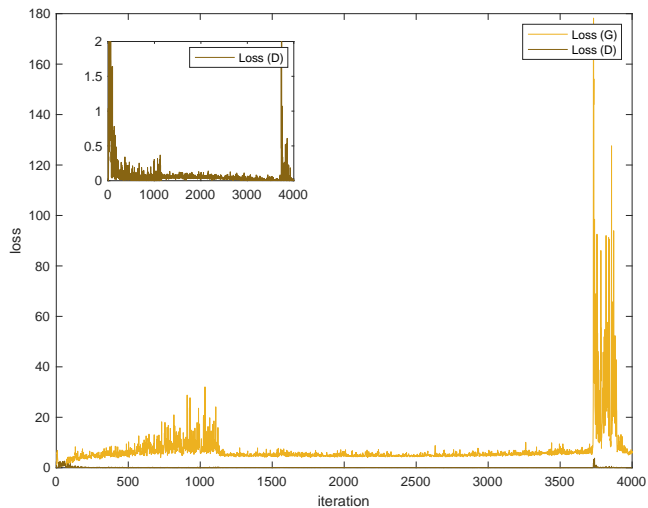}
	\caption{Training set losses convergence by GANs with $n=128$ and $K=128$ pre-trained inputs.}
    \label{f.train_loss_128}
\end{figure}

Generative Adversarial Networks are known for their generation capabilities, and even though they are capable of providing interesting results, they are still black-boxes concerning their discriminators' and generators' stability. A practical tool is to inspect their convergence visually and create insights according to their outputs. Thus, we opted to select two contrasting experiments to fulfill the analysis mentioned above, such that Figures~\ref{f.train_loss_32} and~\ref{f.train_loss_128} illustrate the training set losses convergence by GANs with $n=32$ and $n=128$, respectively, using $K=128$ pre-trained inputs.

Glancing at Figure~\ref{f.train_loss_32}, one can perceive that even though the discriminator smoothly converged, the generator suffered from spiking values through all the training procedure. Such a fact is explained due to the small number of hidden units used to create the pre-trained RBMs, resulting in smaller-sized data fed to the GANs and hence, being incapable of being adequately sampled with large numbers of noise dimensions ($n_z=10000$). On the other hand, Figure~\ref{f.train_loss_128} illustrates a satisfactory training convergence, where both discriminator and generator suffered from spiking losses in their early and last iterations and attained stable values during the rest of the training. Such behavior reflects in its performance, where $S^{\ast}$ with $n=128$ and $K=128$ achieved a mean reconstruction error of $64.756$ against $88.607$ from $S^{\ast}$ with $n=32$ and $K=128$.

\subsubsection{Influence of $K$ Pre-Trained RBMs}
\label{sss.experiments_rec_n_rbms}

\begin{figure}[!ht]
	\centering
	\includegraphics[scale=0.5]{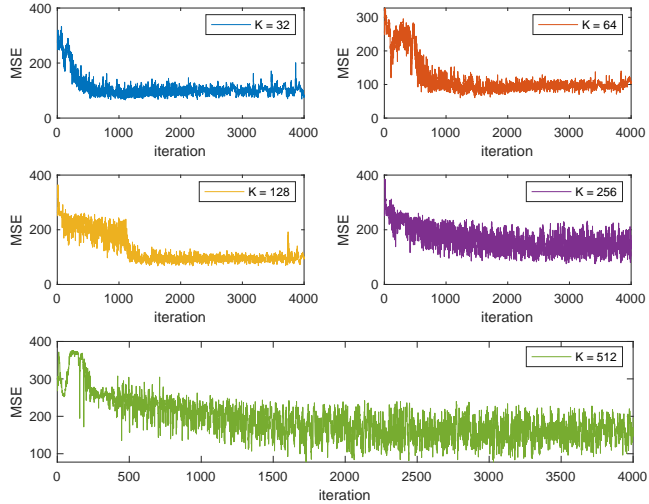}
	\caption{Validation set MSE convergence by GANs with $n=128$ pre-trained $K$ inputs.}
    \label{f.val_rec_128}
\end{figure}

To provide a more in-depth analysis of how the number of pre-trained RBMs influences the experimentation, we opted to plot a convergence graph concerning the validation set reconstruction error obtained by GANs with $n=128$, as illustrated by Figure~\ref{f.val_rec_128}. Such a figure exemplifies the difficulty in training GANs, where an instability often accompanies their training. Moreover, larger $K$ values, such as $K=256$ and $K=512$, brought more instability to the learning process, being visually perceptible in the reconstruction plots' valleys and ridges. On the other hand, a smaller $K$ value, such as $K=32$, did not attain much instability and did not minimize the reconstruction error as $K=64$ and $K=128$.

\subsubsection{Quality of Reconstruction}
\label{sss.experiments_rec_quality}

\begin{figure}[!ht]
	\centering
	\begin{tabular}{cc}
	\includegraphics[scale=0.25]{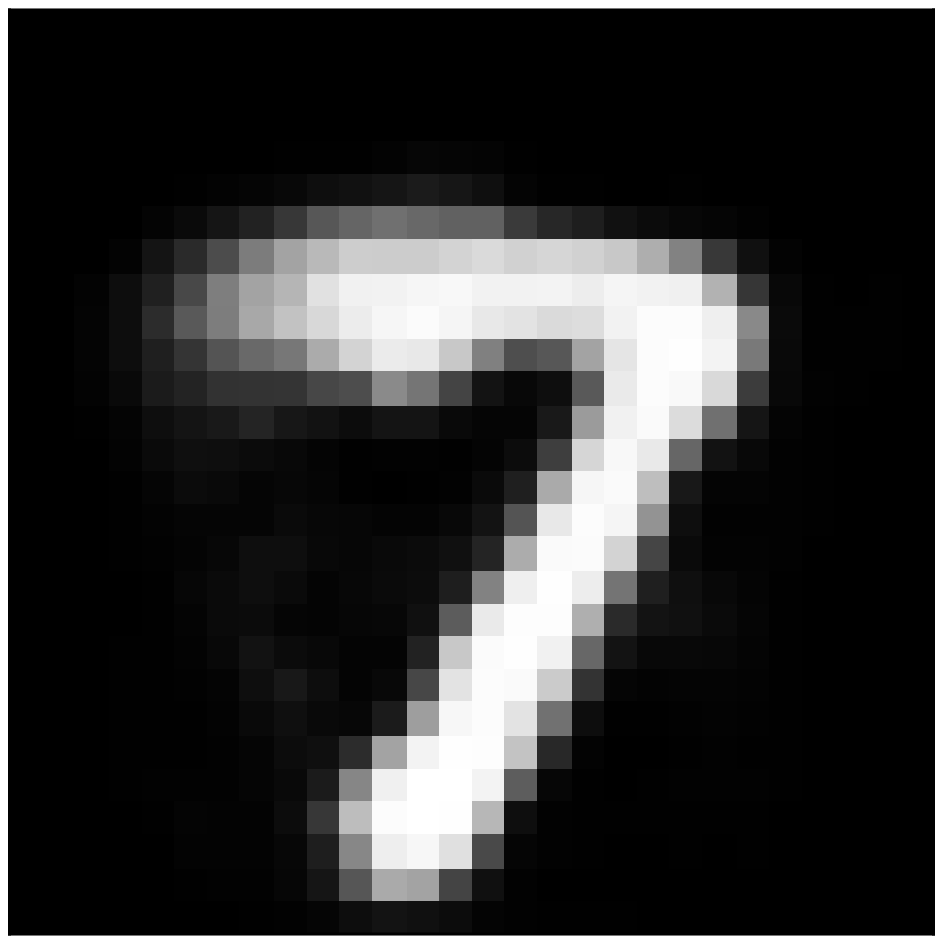} & \includegraphics[scale=0.25]{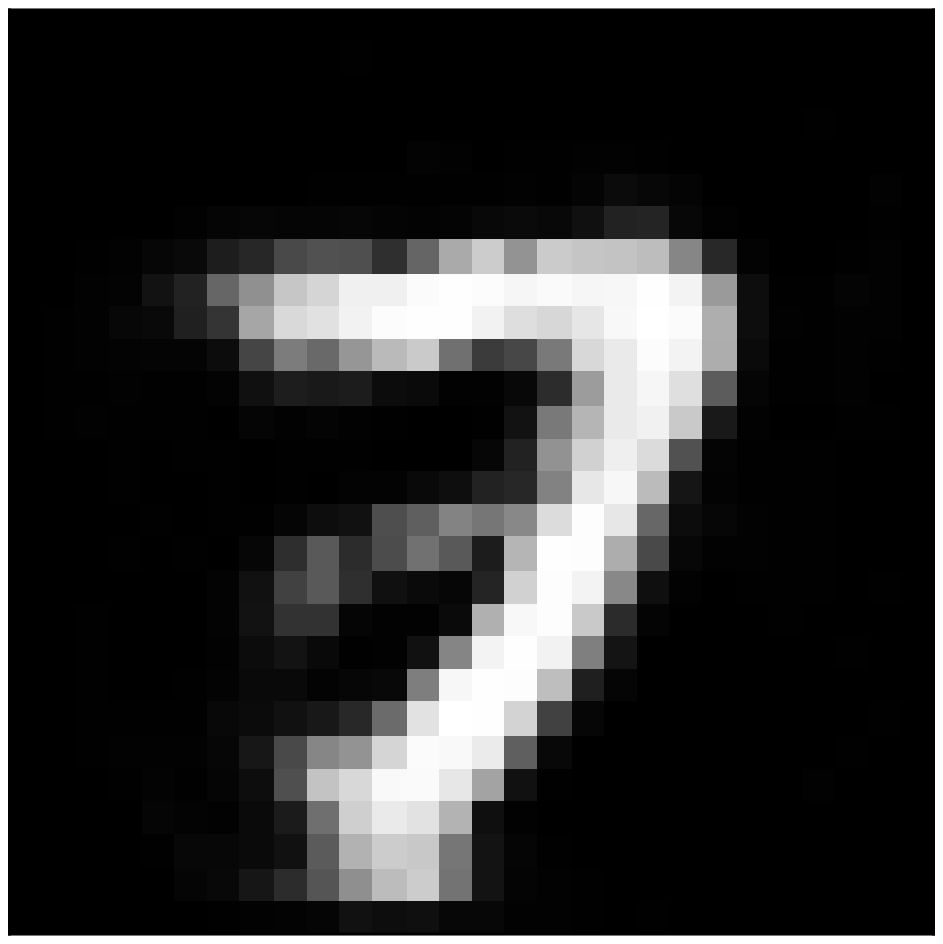} \\
	(a) & (b) \\
	\includegraphics[scale=0.25]{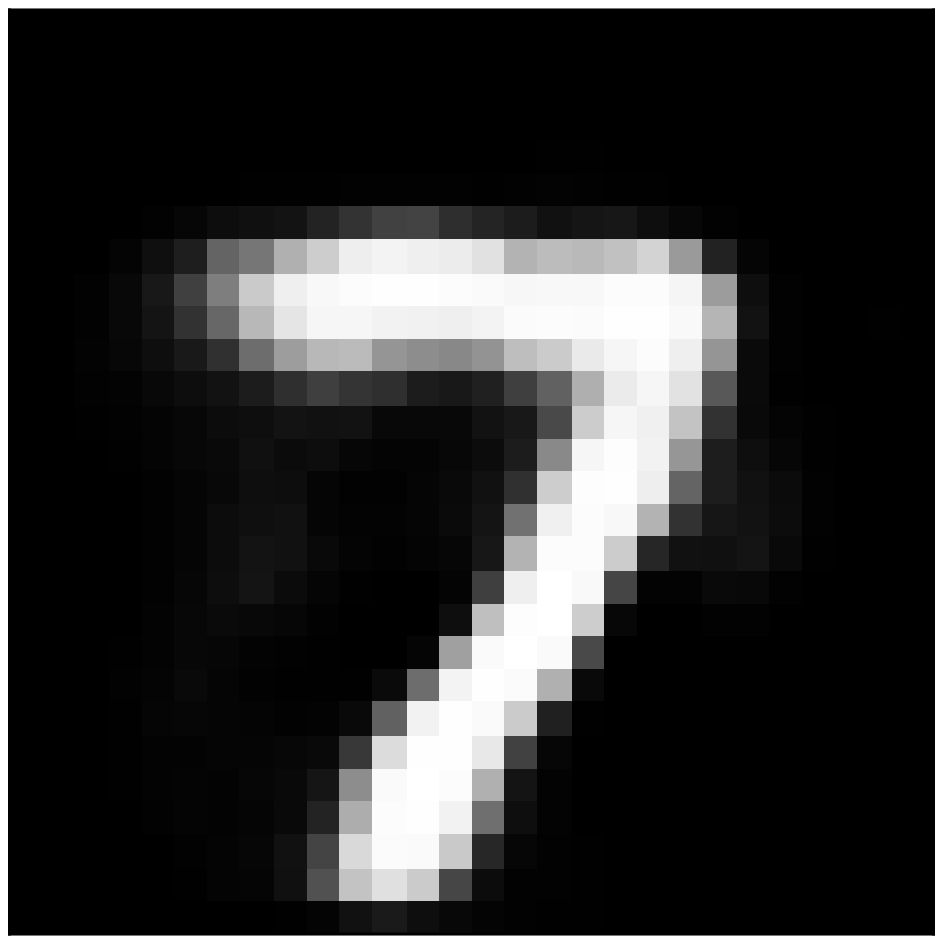} & \includegraphics[scale=0.25]{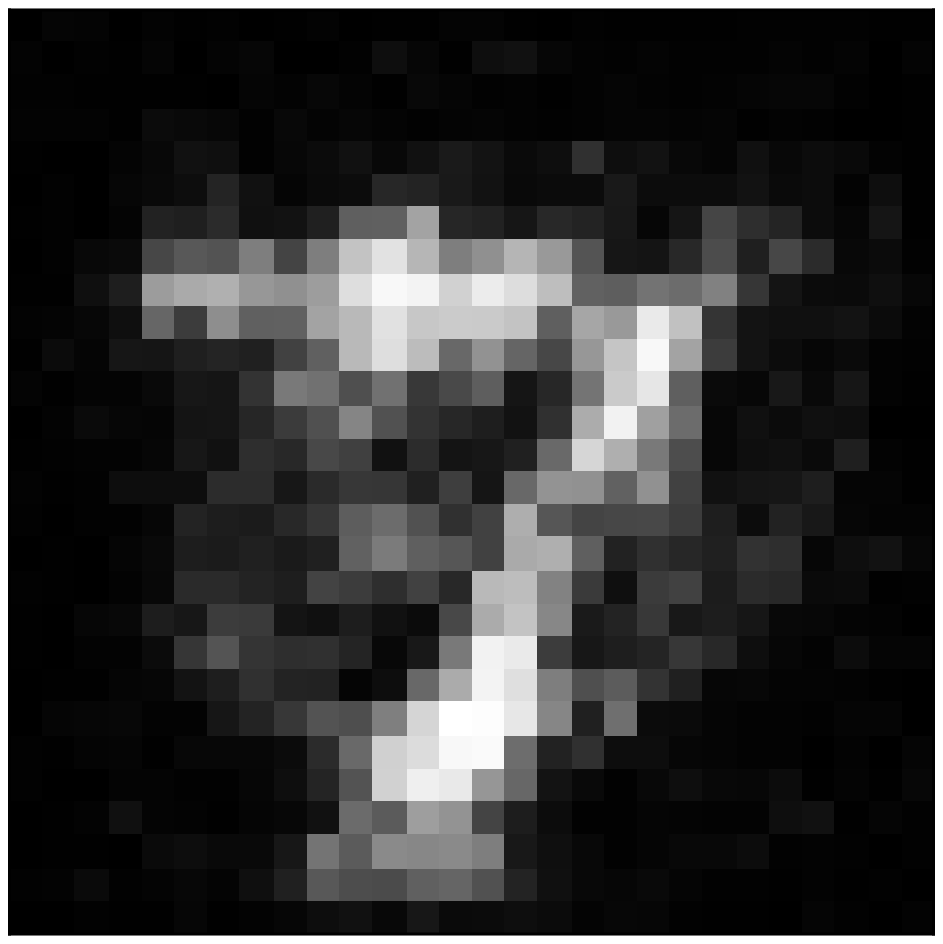} \\
	(c) & (d)
	\end{tabular}
	\caption{Random sample reconstructed by (a) RBM and (b) GAN using $n=128$ and $K=128$, while (c) and (d) stand for RBM and (b) GAN using $n=128$ and $K=512$, respectively.}
    \label{f.quality_rec}
\end{figure}

When analyzing a network's reconstruction capacity, it is common to analyze a sample of a reconstructed image. Figure~\ref{f.quality_rec} illustrates a random reconstructed sampled obtained by an RBM and a GAN using $n=128$ and $K=128$, as well as an RBM and a GAN using $n=128$ and $K=512$. One can perceive that the original RBM reconstructed samples, (a) and (c), were the most pleasant ones, as expected due to their lower MSE. Furthermore, the closest MSE an artificial RBM could achieve compared to the traditional RBMs is depicted by (c), where it could almost reconstruct the sample in the same way as the original versions did. On the other hand, the worst artificial RBM is illustrated by (d), showing that when a learning procedure carries too much noise, it often does not learn the correct patterns that it was supposed to learn.

\subsection{Image Classification}
\label{ss.experiments_clf}

Regarding the image classification task, we opted to employ an additional set of ensemble architectures, where $H_{R}$ stands for an ensemble composed of only original RBMs, while $H_{S^{\ast}}$ stands for an artificial-based ensemble. Additionally, we proposed the following heterogeneous ensembles:

\begin{itemize}
\item $H_{\alpha}$: composed by $0.5K$ original and $0.5K$ artificial RBMs;
\item $H_{\beta}$: composed by $0.25K$ original and $0.75K$ artificial RBMs;
\item $H_{\gamma}$: composed by $K$ original and $5K$ artificial RBMs.
\end{itemize}

Table~\ref{t.rbm_clf_exp} describes the mean classification accuracies and their standard deviation over the MNIST testing set. According to the Wilcoxon signed-rank test with $5\%$ significance, the bolded cells are statistically equivalent, and the underlined ones are the highest mean classification accuracies.

\begin{table}[!ht]
	\renewcommand{\arraystretch}{1.25}
	\caption{Mean classification accuracy (\%) and standard deviation over MNIST testing set.}
	\label{t.rbm_clf_exp}
	\centering
	\scalebox{0.8}{
	\hspace*{-4cm}
	\begin{tabular}{ccccccccc}
	\toprule
	\textbf{Hidden Units} & \textbf{Number of RBMs} & $\mathbf{R}$ & $\mathbf{S^{\ast}}$ & $\mathbf{H_R}$ & $\mathbf{H_{S^{\ast}}}$ & $\mathbf{H_{\alpha}}$ & $\mathbf{H_{\beta}}$ & $\mathbf{H_{\gamma}}$ \\
	\midrule
	\multirow{5}{*}{$32$} & $32$ & $90.94 \pm 0.45$ & $91.68 \pm 0.22$ & $\underline{\mathbf{94.44 \pm 0.00}}$ & $91.98 \pm 0.00$ & $92.95 \pm 0.00$ & $92.23 \pm 0.00$ & $92.29 \pm 0.00$ \\
	& $64$ & $90.88 \pm 0.49$ & $86.45 \pm 0.21$ & $\underline{\mathbf{94.61 \pm 0.00}}$ & $86.75 \pm 0.00$ & $90.93 \pm 0.00$ & $87.70 \pm 0.00$ & $87.23 \pm 0.00$ \\
	& $128$ & $90.86 \pm 0.46$ & $86.69 \pm 0.70$ & $\underline{\mathbf{94.70 \pm 0.00}}$ & $87.89 \pm 0.00$ & $91.96 \pm 0.00$ & $89.07 \pm 0.00$ & $88.60 \pm 0.00$ \\
	& $256$ & $90.88 \pm 0.43$ & $83.34 \pm 0.71$ & $\underline{\mathbf{94.67 \pm 0.00}}$ & $85.54 \pm 0.00$ & $91.78 \pm 0.00$ & $87.76 \pm 0.00$ & $86.83 \pm 0.00$ \\
	& $512$ & $90.90 \pm 0.41$ & $85.19 \pm 1.03$ & $\underline{\mathbf{94.73 \pm 0.00}}$ & $88.03 \pm 0.00$ & $92.59 \pm 0.00$ & $90.04 \pm 0.00$ & $89.13 \pm 0.00$ \\
	\midrule
	\multirow{5}{*}{$64$} & $32$ & $94.17 \pm 0.22$ & $93.86 \pm 0.28$ & $\underline{\mathbf{95.83 \pm 0.00}}$ & $93.96 \pm 0.00$ & $94.88 \pm 0.00$ & $94.16 \pm 0.00$ & $94.01 \pm 0.00$ \\
	& $64$ & $94.17 \pm 0.21$ & $92.88 \pm 0.27$ & $\underline{\mathbf{95.82 \pm 0.00}}$ & $93.21 \pm 0.00$ & $94.47 \pm 0.00$ & $93.67 \pm 0.00$ & $93.47 \pm 0.00$ \\
	& $128$ & $94.18 \pm 0.21$ & $94.41 \pm 0.14$ & $\underline{\mathbf{95.78 \pm 0.00}}$ & $94.60 \pm 0.00$ & $95.07 \pm 0.00$ & $94.72 \pm 0.00$ & $94.71 \pm 0.00$ \\
	& $256$ & $94.19 \pm 0.21$ & $91.89 \pm 0.53$ & $\underline{\mathbf{95.88 \pm 0.00}}$ & $92.68 \pm 0.00$ & $94.56 \pm 0.00$ & $93.36 \pm 0.00$ & $93.06 \pm 0.00$ \\
	& $512$ & $94.18 \pm 0.21$ & $92.12 \pm 0.30$ & $\underline{\mathbf{95.85 \pm 0.00}}$ & $93.25 \pm 0.00$ & $94.98 \pm 0.00$ & $93.92 \pm 0.00$ & $93.67 \pm 0.00$ \\
	\midrule
	\multirow{5}{*}{$128$} & $32$ & $95.39 \pm 0.15$ & $95.21 \pm 0.08$ & $\underline{\mathbf{96.20 \pm 0.00}}$ & $95.25 \pm 0.00$ & $95.47 \pm 0.00$ & $95.39 \pm 0.00$ & $95.30 \pm 0.00$ \\
	& $64$ & $95.37 \pm 0.13$ & $95.26 \pm 0.08$ & $\underline{\mathbf{96.24 \pm 0.00}}$ & $95.33 \pm 0.00$ & $95.61 \pm 0.00$ & $95.41 \pm 0.00$ & $95.35 \pm 0.00$ \\
	& $128$ & $95.37 \pm 0.15$ & $95.42 \pm 0.07$ & $\underline{\mathbf{96.23 \pm 0.00}}$ & $95.54 \pm 0.00$ & $95.82 \pm 0.00$ & $95.66 \pm 0.00$ & $95.62 \pm 0.00$ \\
	& $256$ & $95.37 \pm 0.16$ & $94.34 \pm 0.19$ & $\underline{\mathbf{96.27 \pm 0.00}}$ & $94.79 \pm 0.00$ & $95.85 \pm 0.00$ & $95.15 \pm 0.00$ & $95.02 \pm 0.00$ \\
	& $512$ & $95.37 \pm 0.15$ & $93.84 \pm 0.26$ & $\underline{\mathbf{96.26 \pm 0.00}}$ & $94.35 \pm 0.00$ & $95.49 \pm 0.00$ & $94.72 \pm 0.00$ & $94.65 \pm 0.00$ \\
	\bottomrule
	\end{tabular}}
\end{table}

According to Table~\ref{t.rbm_clf_exp}, it is possible to visualize that standard RBMs $R$ could outperform artificial RBMs $S^{\ast}$ in $13$ out of $15$ experiments, except when $n=32$ and $K=32$, $n=64$ and $K=128$, and $n=128$ and $K=128$. On the other hand, standard ensemble $H_R$ outperformed artificial ensemble $H_{S^{\ast}}$ in every experiment, attaining the highest mean accuracies. An interesting fact is that the accuracy gap between both $H_R$ and $H_{S^{\ast}}$ ensembles has decreased as the number of hidden units has increased. Additionally, it is possible to observe that a high number of $K$ did not impact $H_R$ as much $H_{S^{\ast}}$ has been impacted, mainly because the noise obtained from $S^{\ast}$ has been propagated to their ensembles.

When comparing the heterogeneous ensembles (mix of standard and artificial RBMs), one can perceive that fewer artificial RBMs improved the performance of the ensemble, where the best one $H_{\alpha}$ is only composed of $50\%$ artificial RBMs. An elevated number of artificial RBMs, depicted in $H_{\beta}$ and $H_{\gamma}$, highly degraded their ensembles performance, yet it was attenuated as the number of hidden units was increased. Furthermore, an excessive number of artificial RBMs ($5K$) has not degraded much performance when applied in conjunction with every original RBM ($K$), depicted by the $H_{\gamma}$ column.

\subsection{Complexity Analysis}
\label{ss.experiments_clf_complexity}

Let $\alpha$ be the complexity of pre-training an RBM, while $\beta$ and $\gamma$ be the training and validation complexities of a GAN, respectively. It is possible to observe that $K$ RBMs were pre-trained and fed to the GAN learning procedure. Hence, the whole learning procedure complexity is depicted by Equation~\ref{e.train_complexity}, as follows:

\begin{equation}
\label{e.train_complexity}
\alpha \cdot K + \beta + \gamma.
\end{equation}

The proposed approach intends to sample new artificial RBMs from a pre-trained GAN and compare them against the standard pre-trained RBMs. Let $\iota$ be the complexity of sampling $L$ RBMs, and the whole sampling procedure complexity described by Equation~\ref{e.sampling_complexity}, as follows:

\begin{equation}
\label{e.sampling_complexity}
\iota \cdot L
\end{equation}

Therefore, summing both Equations~\ref{e.train_complexity} and~\ref{e.sampling_complexity} together, it is possible to achieve the proposed approach complexity, as described by Equation~\ref{e.total_complexity}:

\begin{equation}
\label{e.total_complexity}
\alpha \cdot K + \beta + \gamma + \iota \cdot L.
\end{equation}

Note that $\iota \cdot L$ is insignificant compared to $\alpha \cdot K$, even when $L$ is significantly larger than $K$. The advantage of the proposed approach happens when $\beta + \gamma$ is comparable to $\alpha \cdot K$, i.e., when $K$ tends to a large number, the cost of training and validating GANs will be smaller than training new sets of RBMs, thus sampling new RBMs from GANs will be less expensive that training new RBMs.

%% file: conclusion.tex
\section{Conclusion}
\label{s.conclusion}

This work addressed the generation of artificial Restricted Boltzmann Machines through Generative Adversarial Networks. The overall idea was to alleviate the burden of pre-training vast amounts of RBMs by learning a GAN-based latent space representing the RBMs' weight matrices, where new matrices could be sampled from a random noise input. Additionally, such an approach was employed to construct heterogeneous ensembles and further improve standard- and artificial-based RBMs' recognition ability.

The presented experimental results were evaluated over a classic literature dataset, known as MNIST, and conducted over two different tasks: image reconstruction and image classification. Considering the former task, RBMs were trained to minimize the reconstruction error while the latter task uses RBMs as feature extractors and an additional Support Vector Machine to perform the classification.

Considering the image reconstruction task, original RBMs were capable of reconstructing images better in all the evaluated configurations, yet in some cases, artificial RBMs were able almost to reconstruct as equal. A thorough analysis between the number of hidden units ($n$) and the number of sampled RBMs ($K$) depicted that artificial-based RBMs could reconstruct better when $n$ was the highest possible and $K$ was not high nor low. Such behavior endures in the GAN training procedure, where the fewer amount of hidden units resulted in smaller-sized data and could not accurately be sampled with a high number of noise dimensions. Additionally, a large amount of $K$ brought instability to the learning procedure as it increased the amount of diversity and caused the loss function to spike (creation of several valleys and ridges) during the training epochs.

Regarding the image classification task, ensembles with original RBMs could classify the employed dataset better. Nevertheless, it is essential to remark that a high number of hidden units decreased the gap between original and artificial RBMs classification, e.g., $95.39\% \times 95.21\%$ with $n=128$ and $K=32$, $95.37\% \times 95.26\%$ with $n=128$ and $K=64$, and $95.37\% \times 95.42\%$ with $n=128$ and $K=128$. On the other hand, one can perceive that a high number of artificial RBMs in the ensembles brought a decrease in performance, while fewer artificial RBMs in the ensembles attained the second-best performance. Such behavior is explained by the base models $R$ and $S^{\ast}$, where $R$ was better in $12$ out of $15$ experiments, thus impacting more in the ensembles' performance.

For future works, we aim to expand the proposed approach to more challenging image datasets, such as CIFAR and Caltech101, as well as employ it in text datasets, such as IMDB and SST. Moreover, we aim at increasing the number of hidden units and sampled RBMs to verify whether the improvement trend continues to happen. Finally, we intend to use more complex GAN-based architectures, such as DCGAN and WGAN, to overcome its instability complication.

%% file: paper.bbl
\begin{thebibliography}{10}
\expandafter\ifx\csname url\endcsname\relax
  \def\url#1{\texttt{#1}}\fi
\expandafter\ifx\csname urlprefix\endcsname\relax\def\urlprefix{URL }\fi
\expandafter\ifx\csname href\endcsname\relax
  \def\href#1#2{#2} \def\path#1{#1}\fi

\bibitem{Bishop:06}
C.~Bishop, Pattern recognition and machine learning, springer, 2006.

\bibitem{Wang:17}
J.~Wang, L.~Perez, The effectiveness of data augmentation in image
  classification using deep learning, Convolutional Neural Networks Vis.
  Recognit (2017) 11.

\bibitem{Wang:16}
X.~Wang, et~al., Deep learning in object recognition, detection, and
  segmentation, Foundations and Trends{\textregistered} in Signal Processing
  8~(4) (2016) 217--382.

\bibitem{Sallab:17}
A.~Sallab, M.~Abdou, E.~Perot, S.~Yogamani, Deep reinforcement learning
  framework for autonomous driving, Electronic Imaging 2017~(19) (2017) 70--76.

\bibitem{Deng:18}
L.~L.~Deng, Y.~Liu, Deep learning in natural language processing, Springer,
  2018.

\bibitem{Hinton:02}
G.~Hinton, Training products of experts by minimizing contrastive divergence,
  Neural computation 14~(8) (2002) 1771--1800.

\bibitem{Srivastava:12}
N.~Srivastava, R.~Salakhutdinov, Multimodal learning with deep boltzmann
  machines, in: Advances in neural information processing systems, 2012, pp.
  2222--2230.

\bibitem{Szegedy:13}
C.~Szegedy, W.~Zaremba, I.~Sutskever, J.~Bruna, D.~Erhan, I.~Goodfellow,
  R.~Fergus, Intriguing properties of neural networks, arXiv preprint
  arXiv:1312.6199.

\bibitem{Laskov:10}
P.~Laskov, R.~Lippmann, Machine learning in adversarial environments (2010).

\bibitem{Li:17}
C.~Li, H.~Liu, C.~Chen, Y.~Pu, L.~Chen, R.~Henao, L.~Carin, Alice: Towards
  understanding adversarial learning for joint distribution matching, in:
  Advances in Neural Information Processing Systems, 2017, pp. 5495--5503.

\bibitem{Goodfellow:14}
I.~Goodfellow, J.~Pouget-Abadie, M.~Mirza, B.~Xu, D.~Warde-Farley, S.~Ozair,
  A.~Courville, Y.~Bengio, Generative adversarial nets, in: Advances in neural
  information processing systems, 2014, pp. 2672--2680.

\bibitem{Ledig:17}
C.~Ledig, L.~Theis, F.~Husz{\'a}r, J.~Caballero, A.~Cunningham, A.~Acosta,
  A.~Aitken, A.~Tejani, J.~Totz, Z.~Wang, et~al., Photo-realistic single image
  super-resolution using a generative adversarial network, in: Proceedings of
  the IEEE conference on computer vision and pattern recognition, 2017, pp.
  4681--4690.

\bibitem{Choi:18}
Y.~Choi, M.~Choi, M.~Kim, J.-W. Ha, S.~Kim, J.~Choo, Stargan: Unified
  generative adversarial networks for multi-domain image-to-image translation,
  in: Proceedings of the IEEE conference on computer vision and pattern
  recognition, 2018, pp. 8789--8797.

\bibitem{Ma:19}
J.~Ma, W.~Yu, P.~Liang, C.~Li, J.~Jiang, Fusiongan: A generative adversarial
  network for infrared and visible image fusion, Information Fusion 48 (2019)
  11--26.

\bibitem{Zhao:16}
J.~Zhao, M.~Mathieu, Y.~LeCun, Energy-based generative adversarial network,
  arXiv preprint arXiv:1609.03126.

\bibitem{Luo:18}
L.~Luo, S.~Zhang, Y.~Wang, H.~Peng, An alternate method between generative
  objective and discriminative objective in training classification restricted
  boltzmann machine, Knowledge-Based Systems 144 (2018) 144--152.

\bibitem{Zhang:19}
J.~Zhang, S.~Ding, N.~Zhang, W.~Jia, Adversarial training methods for boltzmann
  machines, IEEE Access.

\bibitem{Hinton:12}
G.~Hinton, A practical guide to training restricted boltzmann machines, in:
  G.~Montavon, G.~Orr, K.-R. M{\"u}ller (Eds.), Neural Networks: Tricks of the
  Trade, Vol. 7700 of Lecture Notes in Computer Science, Springer Berlin
  Heidelberg, 2012, pp. 599--619.

\bibitem{Chang:11}
C.-C. Chang, C.-J. Lin, Libsvm: A library for support vector machines, ACM
  transactions on intelligent systems and technology (TIST) 2~(3) (2011) 1--27.

\bibitem{Larochelle:08}
H.~Larochelle, Y.~Bengio, Classification using discriminative restricted
  boltzmann machines, in: Proceedings of the 25th international conference on
  Machine learning, 2008, pp. 536--543.

\bibitem{Lecun:98}
Y.~LeCun, L.~Bottou, Y.~Bengio, P.~Haffner, Gradient-based learning applied to
  document recognition, Proceedings of the IEEE 86~(11) (1998) 2278--2324.

\bibitem{Roder:20}
M.~Roder, G.~H. de~Rosa, J.~P. Papa, Learnergy: Energy-based machine learners
  (2020).
\newblock \href {http://arxiv.org/abs/2003.07443} {\path{arXiv:2003.07443}}.

\end{thebibliography}
